\documentclass[conference]{IEEEtran}
\IEEEoverridecommandlockouts
\usepackage{cite}
\usepackage{amsmath,amssymb,amsfonts}
\usepackage{algorithmic}
\usepackage{graphicx}
\usepackage{textcomp}
\usepackage{xcolor}
\def\BibTeX{{\rm B\kern-.05em{\sc i\kern-.025em b}\kern-.08em
    T\kern-.1667em\lower.7ex\hbox{E}\kern-.125emX}}
\usepackage{tabularx}
\usepackage{booktabs}
\usepackage{float}
\usepackage[utf8]{inputenc}
\DeclareUnicodeCharacter{2265}
{$\geq$}
\usepackage{array}
 
\usepackage{placeins}  
\usepackage{float}      

\usepackage{float}  
\usepackage{graphicx}  
\usepackage{caption}  
\usepackage{booktabs}  
\usepackage{multicol}  
\usepackage{multirow}  
\usepackage{placeins}  
\usepackage{array}  
\usepackage{adjustbox} 

\usepackage{forest}

\usepackage{graphicx} 
\usepackage{tikz}
\usetikzlibrary{positioning, calc, shapes, arrows.meta}
\usepackage{float} 
\usepackage{booktabs}

\usepackage{graphicx}
\usepackage{float}      
\usepackage{placeins}   
\usepackage{stfloats}   
\usepackage{caption}
\usepackage{subcaption}
\usepackage{booktabs}
\usepackage{tabularx}
\usepackage{array}
\usepackage[inline]{enumitem}
\usepackage{balance}
\usepackage{url}

\newcolumntype{Y}{>{\raggedright\arraybackslash}X}

\author{
{\rm Giuliano Lorenzoni}\\
glorenzo@uwaterloo.ca\\
University of Waterloo\\
Waterloo, Ontario, Canada

\and
{\rm Paulo Alencar}\\
palencar@uwaterloo.ca\\
University of Waterloo \\
Waterloo, Ontario, Canada

\and
{\rm Donald Cowan}\\
dcowan@uwaterloo.ca\\
University of Waterloo \\
Waterloo, Ontario, Canada
}

\begin{document}



\title{An Agentic LLM-Based Framework for Population-Scale Mental Health Screening}
\maketitle

\begin{abstract}
Mental health disorders affect millions worldwide, and healthcare systems are increasingly overwhelmed by the volume of clinical data generated from electronic records, telemedicine platforms, and population-level screening programs. At the same time, the emergence of novel AI-based approaches in healthcare calls for intelligent frameworks capable of processing domai-specific unstructured clinical information while adapting to patient-specific needs. This paper proposes an agentic framework for building robust LLM-based pipelines, where each stage is encapsulated as a LangChain agent governed by explicit policies and proxy-guided evaluation. Stages are incrementally locked once validated, ensuring that later adaptations cannot overwrite configurations without demonstrated improvement. The proposed framework evolves from feature-level exploration, through proxy-based tuning and freeze/rollback mechanisms, to full orchestration by an Orchestrator Agent that coordinates preprocessing, retrieval, selection, diversity, threshold optimization, and decoding. A proof-of-concept in transcript-based depression detection demonstrates that the framework converges to stable configurations (e.g., cosine similarity, dynamic Top-k, threshold $\tau \approx 0.75$) while controlling evaluation costs and avoiding regressions. These results highlight the potential of agentic AI to enable  population-level mental health screening over large clinical datasets, addressing critical challenges in trustworthiness, reproducibility, and adaptability required in healthcare environments.

\end{abstract}


\section{Introduction}
Mental health has become a major public health priority worldwide, with rising demand for early detection and population-scale screening. The rapid expansion of healthcare data from clinical transcripts, electronic health records, and telehealth platforms has created an environment that requires scalable, intelligent, and adaptive AI systems. However, current machine learning and LLM-based solutions struggle to provide trustworthy, reproducible, and configurable pipelines capable of handling variability across patients, clinical settings, and deployment conditions.

However, the development of agentic-based software frameworks is still in its infancy.  Such development faces several challenges as these frameworks aim to operate more effectively in increasingly complex environments, including the need to understand how to characterize their framework architecture and how such architecture can be configurable to address tasks that require contextual intelligence and adaptability. A specific gap is providing agentic framework topologies with components that have clear roles and responsibilities. Another gap is enabling systems to adjust to evolving scenarios based on the variability of the agentic components.

This paper presents our ongoing work on developing an adaptive agentic-based framework for NLP classification tasks. We define an agentic pipeline for transcript-base classification that includes the definition of core stages and strategies, and a modular agentic execution pipeline. The proposed pipeline involves agent components to support strategies related to (i) tokenization method, (ii) truncation method, (iii) retrieval, (iv) classification, and (v) evaluation metrics. Agentic roles and their responsibilities and subtasks are described. The variability space for an agentic-based text classification system conceptual architecture is characterized in terms of five configuration components.


Although the experimental validation in this work is conducted on the DAIC-WOZ dataset, which is a widely adopted benchmark for clinical depression detection, the proposed framework is architected to operate at population scale. Its agentic orchestration and configuration locking mechanisms could be scaled to support seamless deployment across larger healthcare pipelines, including electronic health records, large-scale screening platforms, and telemedicine infrastructures that process millions of clinical interactions.

In this paper, we introduce an agentic AI framework designed to support large-scale mental health assessment by orchestrating configurable LLM-based stages governed by explicit policies for reproducibility, adaptability, and cost-aware optimization. Unlike traditional monolithic RAG systems, the proposed framework is aligned with challenges in healthcare by enabling population-level screening, dynamic configuration management, and agentic orchestration that preserves non-regression guarantees. This contribution directly addresses emerging challenges including personalization, adaptation, and robustness in AI-driven clinical decision support.

\section{Background \& Related Work} 
%
%
%

\subsection{RAG Pipelines}

Retrieval‑Augmented Generation (RAG) enhances LLMs by integrating external knowledge through a retrieval mechanism. In RAG, a retriever searches a vector‑indexed knowledge base for contextually relevant information, which is then combined with the user’s query and fed into a generator (e.g., a sequence‑to‑sequence model) to produce informed and grounded responses \cite{rag1,rag2}. In short, RAG architectures often follow a four step pipeline: \begin{enumerate*}[label=(\arabic*)] \item query encoding, when user query is encoded in a vector, \item document retrieval, when relevant documents are retrieved from a knowledge base, \item context augmentation, when retrieved documents and the original query are combined, and \item answer generation, when a generative model produces an answer based on the combined input \end{enumerate*}. Recent developments, based on the automation of these steps and integration with LLM frameworks, such as LangChain\footnote{\url{https://www.langchain.com}}, enabled the development of more sophisticated agents capable of web scratching information for planning or execution \cite{raglangchain}.

\subsection{Dynamic Configuration}

The dynamic configuration of a system refers to the capability of a software system to modify, extend, or adapt its configuration settings at runtime without halting execution \cite{dynamicconfiguration1,dynamicconfiguration2}. This feature is particularly important in distributed, large-scale, or fault‑tolerant environments, where stopping a system for changes may be infeasible or cost‑prohibitive. In the context of LLMs and LLM agents, dynamic configuration refers to the ability of an agentic framework to adapt its configuration values, such as truncation size or temperature, in response to shifting operational constraints like accuracy-latency trade-offs \cite{dynamicconfigurationllm1,dynamicconfigurationllm2,dynamicconfigurationllm3}. The study in \cite{dynamictemperature} selects temperature value based on entropy in text based tasks to achieve a more balanced performance in terms of both generation quality and variety.

\subsection{Hyperparameter Search}

Hyperparameter search refers to the process of systematically selecting the optimal configuration of hyperparameters for a machine learning algorithm \cite{hyperparametersearch1,hyperparametersearch2}. Methods range from simple approaches such as grid search\cite{gridsearch1,gridsearch2} and random search\cite{randomsearch1,randomsearch2}, to more sophisticated techniques including Bayesian optimization\cite{bayesianoptimization1,bayesianoptimization2,bayesianoptimization3} or evolutionary algorithms\cite{evolutionary1,evolutionary2}. The study in \cite{hyperparametersearch-classification} reviews the literature on hyperparameter search and experimentally optimizes some hyperparameters for several classification tasks.

\subsection{LangChain and LLM Agents}

LangChain\footnote{\url{https://www.langchain.com}} is an open-source software framework designed to facilitate the construction of large-language-model-based applications by offering a modular, standardized interface for language models, prompt templates, memory, data retrieval, and tool integration. The LangChain framework enables systematic chaining of components such as LLMs, embedding modules, vector stores, prompt templates, and external data sources to streamline the development of sophisticated, data-aware NLP systems. LangChain supports the development of agentic workflows through tool calling agents, memory and planning integration, and orchestration with LangGraph \footnote{https://www.langchain.com/langgraph}. Langchain-based applications are diverse, ranging from code generation\cite{langchain-codegeneration} tp mental health\cite{langchain-mentalhealth} and education\cite{langchain-education}. The study in \cite{langchain-example} explores how LangChain and LangGraph enable modular agents (e.g., TranslateEnAgent, TranslateFrenchAgent) to collaborate within complex workflows for machine translation while preserving context, scalability, and modularity.

\subsection{LLM Agent Coordination and Infrastructure}

The development of agentic systems, where multiple large language model (LLM) agents collaborate, delegate tasks, or specialize across domains, is still in its infancy. We discuss two main aspects: coordination and infrastructure.

The absence of mature approaches to LLM agent coordination can be attributed to several factors \cite{llm-coordination1,llm-coordination2}. First, LLMs are inherently stochastic, producing outputs that may vary across runs and contexts. This unpredictability complicates the formulation of stable coordination strategies, since agent interactions can yield non-deterministic or compounding errors. Second, there is a lack of formal theoretical foundations for multi-agent LLM systems. Traditional multi-agent systems research in artificial intelligence has emphasized explicit reasoning models, negotiation protocols, and game-theoretic coordination. However, these principles have not yet been robustly adapted to LLM-driven agents, which reason implicitly through natural language rather than through symbolic or explicitly encoded strategies.

In software engineering, the deployment of LLM-based systems also suffers from the absence of systematic policies for hyperparameter selection and configuration management\cite{llm-infrastructure3}. Hyperparameters, including model temperature, context window size, retrieval strategies, or memory configurations, have large impact on system performance, reliability, and cost. Unlike traditional machine learning domains where hyperparameter optimization is well studied \cite{hyperparametersearch1}, the rapidly evolving ecosystem of LLMs has not yet converged on reproducible best practices. Beyond conceptual and methodological gaps, the practical infrastructure required to deploy, manage, and scale LLM-driven systems is still under development, limiting scalability and robustness. High computational costs, fragile multi-step workflows, and the absence of standardized monitoring make coordination difficult. Deployment heterogeneity across cloud, on-premise, and edge environments further complicates reproducibility and portability \cite{llm-infrastructure1,llm-infrastructure2}.

\section{Agentic Architecture for Configuration Selection}

Agentic systems are emerging as a new paradigm in software engineering, where pipelines are structured as collections of agents—each encapsulating a well-defined role, governed by policies, and coordinated by an orchestrator. These systems often rely on large language models (LLMs) to automate complex tasks, ranging from text classification and retrieval to reasoning and decision support. Application domains already include software development, healthcare, customer service, and enterprise management, where the demand for adaptability and robustness is growing rapidly.

Within this broader context, text classification provides a representative and high-impact use case. It spans multiple domains such as spam detection, customer sentiment analysis, news categorization, and fraud monitoring. In healthcare, for example, transcript-based classification can be used to support early detection of depression, enabling scalable and cost-effective clinical screening.

Despite these opportunities, the development of agentic software frameworks remains in its early stages. Key challenges include: (i) how to characterize a pipeline architecture in terms of agent roles and responsibilities; (ii) how to manage variability across multiple configuration dimensions (e.g., embeddings, truncation strategies, retrieval metrics, decoding parameters); and (iii) how to enforce non-regression policies so that adaptations at later stages do not compromise earlier validated configurations. Existing systems often treat these choices as monolithic or ad hoc, which results in fragile and costly solutions.

This paper presents our ongoing work on developing an agentic framework for NLP classification pipelines, instantiated in the task of transcript-based depression detection. Our framework defines nine agents, spanning preprocessing, similarity, retrieval, diversity, post-filters, data expansion, threshold tuning, decoding, and orchestration. Each agent is governed by explicit policies that decide when to adopt, freeze, or rollback a configuration, with the Orchestrator Agent coordinating decisions and enforcing non-regression across the pipeline. Proxy evaluations are employed to reduce the cost of exploration, reserving expensive gold evaluations for promising candidates.

The contributions of this work are threefold: 
(1) we define an agentic architecture where each stage of the pipeline is encapsulated as a LangChain agent with explicit policies and responsibilities; 
(2) we characterize the variability space of text classification pipelines in terms of configuration components and their locking mechanisms; and 
(3) we demonstrate, through a case study in depression detection, how the framework converges to stable configurations while reducing regression risk and controlling evaluation costs. By treating configurability, variability management, and policy enforcement as first-class software engineering concerns, this work advances the state of the art in agentic software systems and contributes to the growing research agenda of agentic software engineering.

\subsection{Overall Agentic Design}

\begin{figure}[!t]
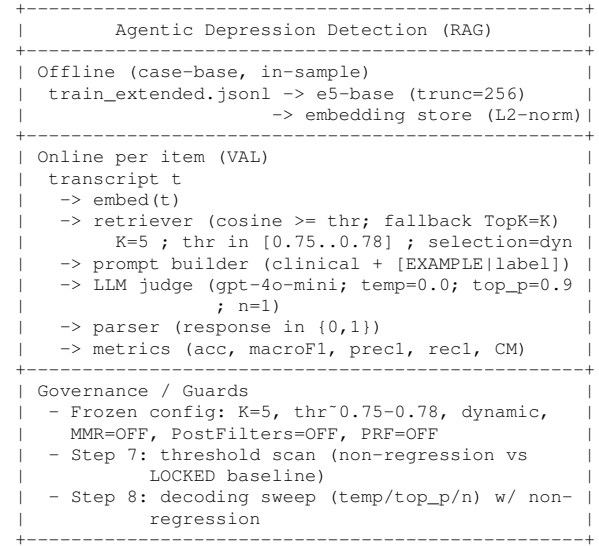

  \centering
  \begin{minipage}{0.85\linewidth}
  \ttfamily\scriptsize
\begin{verbatim}
+--------------------------------------------------+
|        Agentic Depression Detection (RAG)        |
+--------------------------------------------------+
| Offline (case-base, in-sample)                   |
|  train_extended.jsonl -> e5-base (trunc=256)     |
|                      -> embedding store (L2-norm)|
+--------------------------------------------------+
| Online per item (VAL)                            |
|  transcript t                                    |
|   -> embed(t)                                    |
|   -> retriever (cosine >= thr; fallback TopK=K)  |
|        K=5 ; thr in [0.75..0.78] ; selection=dyn |
|   -> prompt builder (clinical + [EXAMPLE|label]) |
|   -> LLM judge (gpt-4o-mini; temp=0.0; top_p=0.9 |
|                 ; n=1)                           |
|   -> parser (response in {0,1})                  |
|   -> metrics (acc, macroF1, prec1, rec1, CM)     |
+--------------------------------------------------+
| Governance / Guards                              |
|  - Frozen config: K=5, thr~0.75-0.78, dynamic,   |
|    MMR=OFF, PostFilters=OFF, PRF=OFF             |
|  - Step 7: threshold scan (non-regression vs     |
|           LOCKED baseline)                       |
|  - Step 8: decoding sweep (temp/top_p/n) w/ non- |
|           regression                             |
+--------------------------------------------------+
\end{verbatim}
  \end{minipage}
  \caption{System view of the agentic RAG pipeline for depression screening.}
  \label{fig:ascii-agentic-rag}
\end{figure}

The proposed agentic pipeline is structured around a system-level view that separates offline case-base construction from online evaluation, as can be seen in Figure~\ref{fig:ascii-agentic-rag}. The design separates the offline case-base construction from the online evaluation path. In the offline stage, transcripts are embedded with a SentenceTransformer and stored in a normalized embedding index. During online execution, each transcript query triggers a retrieval–augmented classification process: embedding, similarity
retrieval with threshold and Top-$k$ fallback, prompt construction with clinical examples, and evaluation by the LLM judge. Final outputs are parsed into binary labels and assessed with accuracy, macro-F1, precision, recall, and confusion matrix. A set of governance guards ensures frozen configurations are not overwritten, enforcing non-regression at decoding time.

\begin{figure}[!t]
    \centering
    \includegraphics[width=\columnwidth]{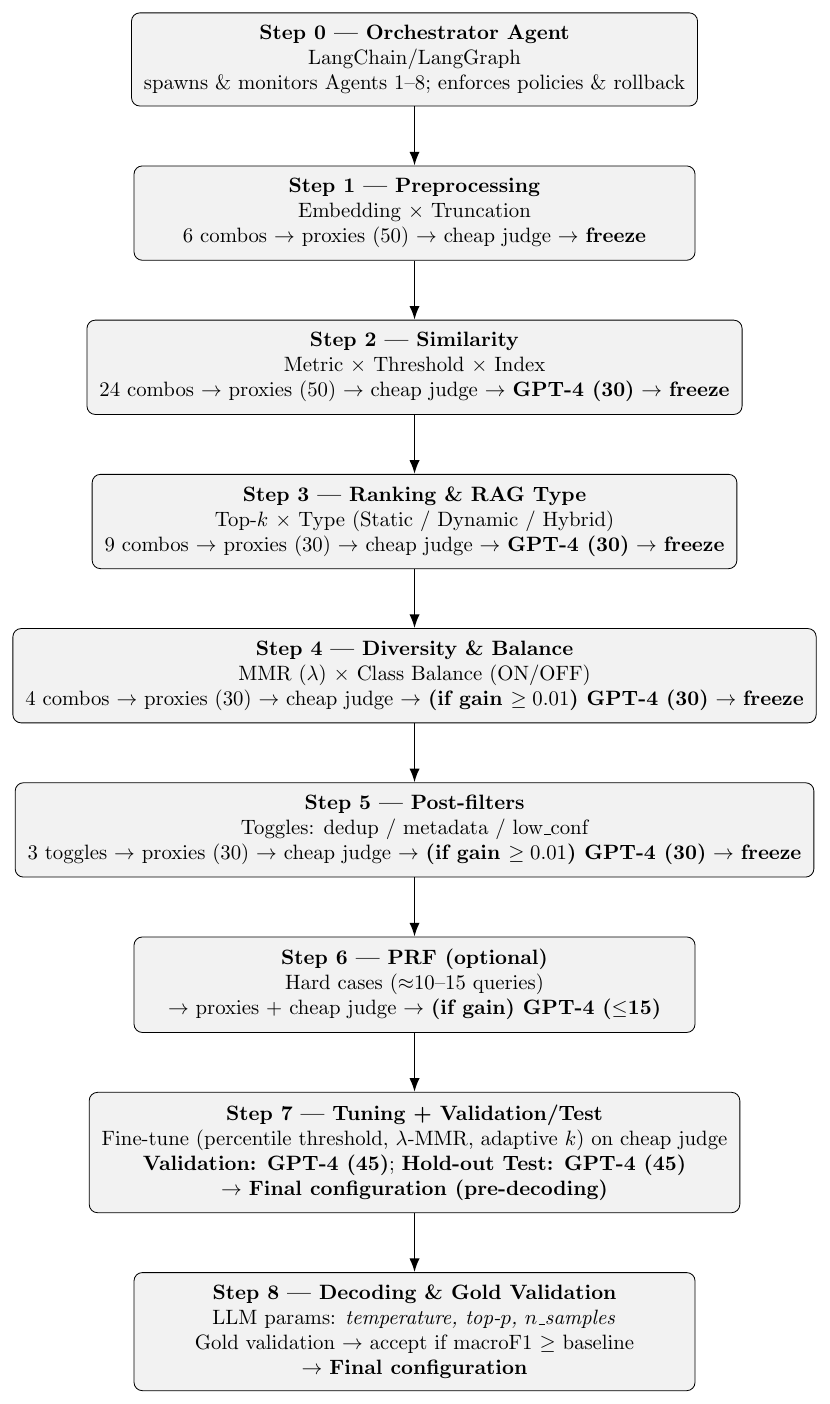}
    \caption{Agentic workflow for transcript-based depression screening. 
    The orchestrator agent (Step~0) coordinates Steps~1–8, enforcing policies and rollback. 
    Each step is executed by a dedicated agent with local policies, while final decoding 
    (temperature, top-$p$, $n$\_samples) and gold validation ensure non-regression.}
    \label{fig:agentic_workflow}
\end{figure}

A complementary workflow-level perspective highlights how the orchestrator coordinates agents across the eight steps while enforcing policies and rollback, as can be seen in Figure~\ref{fig:agentic_workflow}. The orchestrator (Step~0) coordinates specialized agents for preprocessing, retrieval, ranking, diversity, filtering, expansion, threshold optimization, and final validation. Each agent applies local policies, such as freezing or rollback when a regression is detected. This modular architecture supports adaptive configuration while maintaining robustness and cost control. Together, the figures illustrate how our approach combines locked-stage decisions with agentic orchestration to achieve a reliable and extensible software framework.

\subsection{Proxy Analysis}

\begin{table}[!!t]
\caption{Proxy families used to screen configurations before expensive gold validation.}
\label{tab:proxy-families}
\centering
\small
\setlength{\tabcolsep}{4pt}
\renewcommand{\arraystretch}{1.15}
\newcolumntype{Z}{>{\raggedright\arraybackslash}X}
\begin{tabularx}{\linewidth}{@{} l Z Z @{}}
\toprule
\textbf{Proxy family} & \textbf{What it measures} & \textbf{Examples} \\
\midrule
Semantic–retrieval & Quality/coverage of retrieved context & Mean/max similarity, hit@k, coverage of minority class, overlap rate \\
Statistical–text & Text/length/entropy balance & Token count, type–token ratio, stopword ratio, per-class length balance \\
Ranking–stability & Sensitivity to $k$ and re-ranking & Recall vs.\ $k$ curve, Kendall–$\tau$ across rankers, MMR overlap change \\
Confidence heuristics & Heuristic decision certainty & Margin between top similarities, vote entropy of examples \\
Cheap LLM judge & Low-cost LLM agreement & Mini-judge consistency vs.\ gold labels on a small subset \\
\bottomrule
\end{tabularx}
\end{table}

The framework relies heavily on proxies to guide exploration before expensive gold validation. As summarized in Table~\ref{tab:proxy-families}, different proxy families capture complementary aspects of configuration quality, ranging from semantic–retrieval coverage and statistical text balance to ranking stability, confidence heuristics, and lightweight LLM agreement. By combining these low-cost measures, the system can discard unpromising candidates early while retaining diversity among viable options. This reduces evaluation cost and risk of regression, since only configurations that satisfy proxy thresholds are forwarded to gold evaluation. A more detailed mapping of proxies to each agent and step, including non-regression guards and rollback conditions, is provided in the Appendix (Table~\ref{tab:step-proxies-revised}). Together, these mechanisms ensure that proxy-guided screening supports reproducibility and robustness while maintaining the adaptability of the agentic pipeline.

\subsection{Agents and Policies}

\begin{table*}[!t]
\caption{Detailed agents and policies, including freeze/skip criteria and rollback rules.}
\label{tab:agents-policies-appendix}
\centering
\small
\setlength{\tabcolsep}{6pt}
\renewcommand{\arraystretch}{1.2}
\begin{tabularx}{\textwidth}{@{} l X @{}}
\toprule
\textbf{Agent (Step)} & \textbf{Policy (detailed)} \\
\midrule
Step 0 — Orchestrator & Central coordinator. Enforces all policies, prevents overwriting of frozen configs, logs agentic decisions, and triggers automatic rollback if any step leads to regression in macro-F1 or recall. \\
Step 1 — Preprocessing & Test embedding/truncation combos. Freeze only if macro-F1 gain $\geq 0.02$ (proxy or gold). Skip step if all variants regress relative to baseline. \\
Step 2 — Similarity Metric & Default cosine. Switch to alternative metric only if macro-F1 gain $>0.01$. Freeze once locked. Skip metric sweeps if previous frozen config already dominates. \\
Step 3 — Selection (Top-$k$/RAG type) & Increase $k$ until recall stabilizes; enforce $k \leq 5$. Prefer dynamic RAG type. Skip if larger $k$ causes precision collapse or no recall gain. \\
Step 4 — Diversity (MMR) & Activate only if candidate overlap $>0.6$ and no drop in minority recall. Deactivate if macro-F1 falls compared to frozen baseline. \\
Step 5 — Post-filters & Apply filters (dedup/metadata/low\_conf) only if confidence $<0.65$ and no regression in macro-F1. Skip otherwise. \\
Step 6 — Data Expansion & Adopt train+test augmentation only if macro-F1 drop $\leq 0.02$ (non-regression). Freeze expansion if locked config is stable. \\
Step 7 — Threshold Optimization & Sweep around $\tau\approx0.75$; pick best macro-F1 on VAL. Skip if sweep yields no gain $>0.01$. \\
Step 8 — Gold Validation & Accept configuration only if macro-F1 $\geq$ baseline. Reject and rollback otherwise. Freeze final config. \\
\bottomrule
\end{tabularx}
\end{table*}

Each step of the pipeline is implemented as an autonomous agent governed by explicit policies for freezing, skipping, or rolling back configurations. Table~\ref{tab:steps-extended} provides a compact overview of the final locked decisions, while the detailed policies, including non-regression criteria and rollback rules, are listed in the Appendix (Table~\ref{tab:agents-policies-appendix}). These policies ensure that improvements are adopted only when statistically meaningful, and that weaker configurations never overwrite previously frozen baselines. By enforcing such guardrails, the orchestrator (Step~0) coordinates Agents~1–8 in a way that maintains robustness while still allowing controlled exploration of the variability space. This agentic enforcement layer is central to achieving the methodological contribution of the framework: reducing fragility, containing costs, and preserving reproducibility in transcript-based depression detection.


\section{Experimental Use Case: Depression Detection from Transcripts}

\subsection{Dataset}

Our study relies on the Distress Analysis Interview Corpus – Wizard-of-Oz (DAIC-WOZ), a clinical dataset designed to support research on mental health conditions such as depression, anxiety, and post-traumatic stress disorder \cite{10gratch2014distress}. The DAIC-WOZ database is part of a larger corpus and is publicly available to the research community upon request\footnote{\url{https://dcapswoz.ict.usc.edu/}}. In our case, we obtained access to the subset corresponding to patients with depression. 

Due to ethical and legal restrictions, we are not permitted to redistribute the data directly; researchers wishing to replicate our work must submit their own request through the official process. This procedure ensures compliance with privacy regulations while preserving scientific transparency.  

The dataset contains 189 clinical interviews, conducted either by humans, human-controlled agents, or an autonomous animated interviewer called \emph{Ellie}, which was specifically designed to elicit indicators of mental illness. Each session includes multimodal resources such as transcripts, audio, video, and extracted features. The dataset is divided into training, development, and test subsets, and comprises both distressed and non-distressed participants.  

For our classification experiments, we relied on the \texttt{PHQ8\_Score} file, which provides: (i) a participant identifier (\texttt{Participant\_ID}); (ii) a binary flag (\texttt{PHQ8\_Binary}) indicating depression status (0 = depressed, 1 = not depressed); (iii) the numerical severity score (\texttt{PHQ8\_Score}), where values greater than 10 correspond to depressed cases; and (iv) the participant's gender, which was not considered in this study.  

\subsection{Experimental Setup}

\begin{figure}[t]
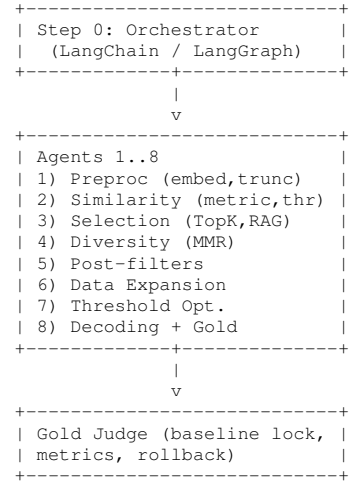

  \centering
  \begin{minipage}{0.55\linewidth}
  \ttfamily\scriptsize
\begin{verbatim}
+----------------------------+
| Step 0: Orchestrator       |
|  (LangChain / LangGraph)   |
+-------------+--------------+
              |
              v
+----------------------------+
| Agents 1..8                |
| 1) Preproc (embed,trunc)   |
| 2) Similarity (metric,thr) |
| 3) Selection (TopK,RAG)    |
| 4) Diversity (MMR)         |
| 5) Post-filters            |
| 6) Data Expansion          |
| 7) Threshold Opt.          |
| 8) Decoding + Gold         |
+-------------+--------------+
              |
              v
+----------------------------+
| Gold Judge (baseline lock, |
| metrics, rollback)         |
+----------------------------+
\end{verbatim}
  \end{minipage}
  \caption{Orchestration overview: Step~0 (LangChain) coordinates Agents~1–8 and validates with the Gold Judge.}
  \label{fig:ascii-orch}
\end{figure}

The experimental setup followed the staged agentic pipeline described in Section~III. 
Each step was executed by a dedicated agent, coordinated by the LangChain-based Orchestrator (Step~0). 
The frozen baseline configuration was established incrementally from Steps~1–6, with non-regression checks applied at each stage, and then refined through threshold optimization (Step~7) and decoding validation (Step~8). 
Evaluation relied on the extended split of the DAIC-WOZ transcripts, where proxies were first used to filter unpromising configurations, and only the most promising variants were validated against the gold judge.
This ensured that the locked pipeline balanced performance (macro-F1 and recall) with efficiency and cost-awareness, while enabling modular analysis of each agentic decision. 
The orchestration of this pipeline is summarized in Figure~\ref{fig:ascii-orch}, which illustrates the interaction between the Orchestrator, Agents~1–8, and the final Gold Judge.

\subsection{Computational Setup}

All experiments were implemented in Python~3.12 with PyTorch 2.8, HuggingFace Transformers 4.55, and Sentence-Transformers 5.1. LangChain (core + OpenAI bindings) was adopted to orchestrate prompt templates, LLM calls, and parsing, providing a declarative layer that materializes the \emph{Agentic Orchestrator} (Step~0). 

Unlike many RAG systems, our pipeline does not require FAISS or heavy vector indices: cosine similarity over NumPy arrays was sufficient, matching the similarity logic from Steps~1–2 and improving re\-producibility. The framework runs comfortably on CPU (Intel i7, 16GB RAM) without GPU acceleration.

Results obtained via LangChain orchestration closely match the ``gold'' evaluations from Steps~7–8, with minor expected variations due to prompt differences or decoding randomness. This validates the agentic pipeline as a software artifact rather than a one-off experiment.







\subsection{Results}


\begin{table*}[t]
\caption{Step-wise results. The base configuration is locked at $K{=}5$, dynamic selection, and MMR~OFF; Steps~7--8 report scans gated by non-regression.}
\label{tab:steps-extended}
\centering
\small
\setlength{\tabcolsep}{6pt}
\renewcommand{\arraystretch}{1.15}
\begin{tabularx}{\textwidth}{@{} l l >{\raggedright\arraybackslash}X l @{}}
\toprule
\textbf{Step} & \textbf{Varied} & \textbf{Key results} & \textbf{Decision} \\
\midrule
1--4 & $K\in\{2,3,5\}$; $\tau\in\{0.75,0.78,0.82\}$; static/dyn; MMR
     & Best proxy: $K=5$, $\tau=0.75$, dynamic; MMR~OFF
     & Lock config \\
5    & Post-filters
     & Baseline (gold, VAL): acc=0.857, macroF1=0.825, rec$_1$=0.875
     & Keep OFF \\
6    & PRF
     & No reliable gain; proxy$+$gold rejected
     & PRF OFF \\
7    & $\tau\in[0.70,0.80]$
     & Peak at $\tau=0.78$ (acc=0.821, macroF1=0.789) \emph{below baseline}
     & \textbf{Reject (non-regression); keep $\tau=0.75$} \\
8    & $\mathrm{temp}\in\{0.0,0.1,0.2\}$,\; $\mathrm{top}\_p\in\{1.0,0.9\}$,\; $n\in\{1,3\}$
     & All settings below baseline; best (0.0, 0.9, 1) = acc=0.821, macroF1=0.789
     & \textbf{No change; keep temp=0.0, top-$p$=1.0, $n$=1} \\
\bottomrule
\end{tabularx}
\end{table*}



Table~\ref{tab:steps-extended} summarizes the outcomes of the step-wise evaluation under the extended split. 
The configuration locked after Steps~1--4 fixed $K{=}5$, dynamic selection, and MMR disabled, as this setting consistently achieved the best proxy balance of recall and stability. 
Subsequent agents confirmed these decisions: Step~5 showed no gain from post-filters, and Step~6 rejected pseudo-relevance feedback due to inconsistent improvements across proxies and gold. 
Step~7 scanned thresholds in the range $\tau\in[0.70,0.80]$, but the apparent peak at $\tau=0.78$ yielded lower accuracy and macro-F1 than the frozen baseline; in line with the non-regression policy, the system retained $\tau=0.75$. 
Similarly, Step~8 evaluated decoding parameters (temperature, top-$p$, $n$), but all variants fell below the locked baseline, leading to no change from the default configuration (temperature$=0.0$, top-$p=1.0$, $n=1$).  

Overall, these results validate the policy-driven orchestration strategy: once a stage was frozen, later agents could only confirm or skip, ensuring non-regression while reducing redundant evaluations. 
The locked pipeline therefore demonstrates that the agentic framework can balance robustness and efficiency while preserving interpretability of each configuration decision.

\section{Discussion}

\subsection{Variability and Non-Regression Guarantees}
One key insight is the variability observed across repeated evaluations of the same configuration. For example, while $\tau{=}0.78$ or decoding with $top\_p{=}0.9$ produced slightly lower scores in a given run, repeated executions of the $\tau{=}0.75$ baseline could in principle yield lower results as well. This stochasticity stems from model sampling effects and the sensitivity of validation splits. The non-regression policy ensures that configurations are not replaced by weaker candidates, but the framework must also tolerate natural variance within expected confidence intervals. This highlights the importance of considering \emph{equivalence classes} of configurations rather than single locked values, and suggests that policy guards should integrate statistical tests or variance-aware thresholds instead of single-run decisions.

\subsection{Adaptability and Variability Space Expansion}
Although our agentic pipeline evaluated a broad set of features, the variability space for each agent is much larger in principle. For example, RAG strategies extend well beyond the static/dynamic/hybrid options tested here, and truncation methods include advanced salience-based or neural summarization approaches not yet explored. The framework must therefore remain \emph{adaptable}: frozen configurations should represent stable checkpoints, but the orchestrator should also support the injection of new candidate policies at runtime. This extensibility is crucial in software engineering practice, where agentic systems must evolve alongside advances in LLM prompting, retrieval, and evaluation strategies.

\subsection{Implications for Agentic SE Systems}
From a broader software engineering perspective, our findings illustrate how agentic pipelines operationalize principles of modularity (e.g., by decomposing
feature engineering and LLM decision-making into a network of agents), rollback, and automated policy enforcement. 
The orchestration logic, implemented via LangChain, demonstrates that agentic SE systems can treat LLM calls as configurable software components rather than opaque black boxes. This architectural stance enables not only reproducibility but also systematic governance, making the framework relevant beyond healthcare to domains such as requirements engineering, code analysis, and risk assessment.

\section{Conclusions and Future Work}



This paper introduced an adaptive agentic framework for transcript-based depression detection, where each configuration step is encapsulated as an autonomous agent governed by explicit policies. By combining proxy-guided exploration with freeze and rollback mechanisms, the framework ensures that once a configuration is locked, later adaptations cannot overwrite it unless they demonstrate meaningful improvements. This design supports robustness, interpretability, and cost-efficiency while mitigating the fragility typically observed in monolithic RAG pipelines.

Our evaluation demonstrated that the framework converges to stable configurations---notably cosine similarity, dynamic Top-$k$, and a frozen threshold at $\tau=0.75$---while discarding unpromising alternatives such as post-filters, PRF, or decoding variants that violated the non-regression policy. These results validate the contribution of agentic enforcement layers: by operationalizing variability management as a first-class concern, the pipeline can remain both auditable and reproducible.

Beyond its technical contributions, this framework establishes a foundational path toward population-scale mental health screening, where agentic AI enables automated, robust, and clinically aligned decision-making. By treating configuration management and policy enforcement as first-class software engineering concerns, the framework can be integrated into digital health infrastructures to support early intervention and large-scale screening initiatives. This cntributes to position agentic AI as a transformative paradigm for future healthcare clinical diagnosis applications.


\balance
\bibliographystyle{IEEEtran}
\bibliography{gpt4_depression}

\clearpage


\section*{\hspace*{-0.1em}APPENDIX}
\addcontentsline{toc}{section}{APPENDIX}

\begin{table*}[!b]
\caption{Mapping of proxies by agent/step}
\label{tab:step-proxies-revised}
\centering
\small
\setlength{\tabcolsep}{6pt}
\renewcommand{\arraystretch}{1.2}
\begin{tabularx}{\textwidth}{@{} l X @{}}
\toprule
\textbf{Agent (Step)} & \textbf{Proxies evaluated (grouped by family)} \\
\midrule
\textbf{Step 0 — Orchestrator} &
\emph{Stability/Seeds}: re-run spot checks to ensure variance within tolerance; 
\emph{Non-regression}: macro-F1/recall never below locked baseline; 
\emph{Cost/Latency}: budget caps per sweep; \emph{Rollback} if violated. \\

\midrule
\textbf{Step 1 — Preprocessing} &
\emph{Semantic similarity}: cosine (query–doc) under truncation modes; 
\emph{Coverage}: salient-term hit-rate (token-256 vs sentence-wise); 
\emph{Length stress}: macro-F1 proxy sensitivity vs truncation length; 
\emph{Stability}: IQR of proxy deltas across 2–3 shuffles. \\

\midrule
\textbf{Step 2 — Similarity Metric} &
\emph{Metric ablation}: cosine vs alternatives (dot, L2-normalized cosine); 
\emph{Ranking (pseudo)}: nDCG@k from similarity order; 
\emph{Minority-recall proxy}: recall@k on positives; 
\emph{Latency}: pairwise time / index-free path confirmed. \\

\midrule
\textbf{Step 3 — Selection (Top-$k$, RAG type)} &
\emph{RAG type}: static vs dynamic vs hybrid; 
\emph{Recall curve}: recall@k for $k\!\in\!\{2,3,5\}$ (stop when saturates); 
\emph{Coverage}: query facet coverage vs $k$; 
\emph{Calibration}: neighbor-label entropy vs $k$ (prefer lower); 
\emph{Context/cost}: average tokens per query; 
\emph{Dynamic gain}: fraction of dynamic replacements and their effect on errors; 
\emph{Non-regression}: macro-F1 proxy $\ge$ baseline. \\

\midrule
\textbf{Step 4 — Diversity (MMR)} &
\emph{Redundancy}: pairwise cosine among top-$k$; 
\emph{Duplicate rate}: near-duplicate fraction ($<\!0.4$ preferred); 
\emph{Novelty ratio}: share of non-overlapping evidence; 
\emph{Minority-recall guard}: no drop when enabling MMR ($\lambda$ sweep); 
\emph{Non-regression}: macro-F1 proxy $\ge$ baseline. \\

\midrule
\textbf{Step 5 — Post-filters} &
\emph{Confidence gate}: apply only if conf.\ $<\!0.65$; 
\emph{Leakage/contamination}: dedup/metadata flags; 
\emph{Safety}: no precision collapse; 
\emph{Non-regression}: macro-F1 proxy not worse than baseline. \\

\midrule
\textbf{Step 6 — Data Expansion (Train\,+\,Test; PRF optional)} &
\emph{Expansion effect}: macro-F1 proxy drift $\le\!0.02$ (non-regression) after adding test items to the retrieval pool; 
\emph{Minority-recall}: change on positives; 
\emph{Leakage guards}: exact ID match and near-dup ($\ge\!0.95$ cosine) across splits; 
\emph{Distribution shift}: vocab/length shift and cosine-centroid shift vs original train; 
\emph{Retrieval behavior}: neighbor-label purity and avg.\ similarity drift; 
\emph{Cost}: tokens/time increase; 
\emph{Decision}: accept only if safe (no-regression) and guards pass. 
\emph{PRF (optional)}: same acceptance rule on a hard-query subset. \\

\midrule
\textbf{Step 7 — Threshold Optimization} &
\emph{Calibration}: sweep around $\tau\!\approx\!0.75$ (e.g., 0.70–0.80) maximizing macro-F1 on VAL; 
\emph{Minority-recall}: prefer operating points preserving depressed recall; 
\emph{Cost}: keep context size reasonable. \\

\midrule
\textbf{Step 8 — Gold Validation / Decoding} &
\emph{Gold judge}: macro-F1, precision/recall and confusion matrix; 
\emph{Decoding sensitivity}: small grid on (temperature, top-$p$, $n$); 
\emph{Non-regression}: accept only if macro-F1 $\ge$ locked baseline; then freeze final config. \\
\bottomrule
\end{tabularx}
\end{table*}

 \end{document}